\documentclass[letterpaper, 10 pt, journal, twoside]{IEEEtran}
\IEEEoverridecommandlockouts
\usepackage{cite}
\usepackage{amsmath,amssymb,amsfonts}
\usepackage{graphicx}
\usepackage{textcomp}
\usepackage{xcolor}
\usepackage{pifont}
\usepackage[english]{babel}
\usepackage[utf8]{inputenc}
\usepackage{algorithm}
\usepackage{cancel}
\usepackage{soul}
\setstcolor{black}
\usepackage{mathtools}
\usepackage{svg}
 
\usepackage{multirow} 
\usepackage[noend]{algpseudocode}
\usepackage{verbatim}
\def\BibTeX{{\rm B\kern-.05em{\sc i\kern-.025em b}\kern-.08em
    T\kern-.1667em\lower.7ex\hbox{E}\kern-.125emX}}
\usepackage[colorlinks=true, allcolors=blue]{hyperref}
\usepackage[normalem]{ulem}
\IEEEpubidadjcol 
\usepackage[top=57pt, left=48pt, right=48pt, bottom=43pt]{geometry}
\def\IEEEtitletopspace{3pt}
\begin{document}
\title{Di-NeRF: Distributed NeRF for Collaborative Learning with Relative Pose Refinement}
\author{Mahboubeh Asadi$^{1}$, Kourosh Zareinia$^{1}$, Sajad Saeedi$^{1}$
\thanks{\textcolor{black}{Manuscript received: August, 03, 2024; Accepted September, 17, 2024.}}
\thanks{\textcolor{black}{This paper was recommended for publication by Editor Javier Civera upon evaluation of the Associate Editor and Reviewers' comments.
This work was supported by the Natural Sciences and
Engineering Research Council of Canada (NSERC).}} 
\thanks{\textcolor{black}{$^{1}$All authors are with 
Toronto Metropolitan University, Canada.}}%
\thanks{\textcolor{black}{Digital Object Identifier (DOI): see top of this page.}}
}
\markboth{IEEE Robotics and Automation Letters. Preprint Version. Accepted September, 2024}
{Asadi \MakeLowercase{\textit{et al.}: Di-NeRF}}
\maketitle
\begin{abstract}
Collaborative mapping of unknown environments can be done faster and more robustly than a single robot. However, a collaborative approach requires a distributed paradigm to be scalable and deal with communication issues. This work presents a fully distributed algorithm enabling a group of robots to collectively optimize the parameters of a Neural Radiance Field (NeRF). The algorithm involves the communication of each robot's trained NeRF parameters over a mesh network, where each robot trains its NeRF and has access to its own visual data only. Additionally, the relative poses of all robots are jointly optimized alongside the model parameters, enabling mapping with \textcolor{black}{less accurate} relative camera poses. We show that multi-robot systems can benefit from differentiable and robust 3D reconstruction optimized from multiple NeRFs. Experiments on real-world and synthetic data demonstrate the efficiency of the proposed algorithm. 
\setcounter{footnote}{1}
See the website of the project for videos of the experiments and supplementary material\footnote{\href{https://sites.google.com/view/di-nerf/home}{https://sites.google.com/view/di-nerf/home}}.

\end{abstract}
\section{Introduction}
\IEEEPARstart{T}{here} is an increasing demand for robots to collaborate on complex tasks, such as mapping an unknown environment. Centralized approaches for the coordination of the robots or processing data face scalability issues, vulnerability to central node failures, and the risk of communication blackouts. A true collaborative robotic system needs to work in a distributed manner~\cite{halsted2021survey}. Further, building a high-quality representation in collaborative mapping is needed to make an informed decision. The representations also need to be compact for easy sharing. Neural Radiance Field (NeRF)~\cite{mildenhall2021nerf} enables such a representation for a single robot, leveraging advancements in neural networks. Extending the capabilities of NeRF to facilitate multi-robot NeRF in a distributed manner emerges as a natural progression. This extension allows for \textcolor{black}{a reliable and fast} collaborative development of high-quality representations of unknown environments \textcolor{black}{(e.g. in search \& rescue, surveillance, and monitoring),} without a central node.
Performing distributed NeRF requires addressing several problems inherent to multi-robot systems~\cite{Saeedi2016jfr}. Firstly, determining the relative poses of robots, preferably without requiring them to rendezvous. This is needed to enable the fusions of the local maps (i.e. maps of individual robots) into a global map. The relative poses can be determined either via line-of-sight rendezvous~\cite{Murai2024TRO}, which necessitates motion coordination, or by identifying overlaps within local maps~\cite{Stathoulopoulos2023ICRA}, 
requiring additional processing. In large-scale applications, avoiding line-of-sight rendezvous is preferred to prevent further planning constraints. The second challenge is deciding what information to share among robots. While sharing unprocessed raw visual data is resource-intensive, 
depending on the architecture, sharing compact neural models is a more viable option. However, sharing maps introduces complexity to the subsequent challenge. The third challenge is merging representations, requiring the development of a method to unify local neural maps into a global map. Merging neural maps is more intricate than merging geometric representations.  
\begin{figure}[t]
\centerline{\includegraphics[scale=0.65]{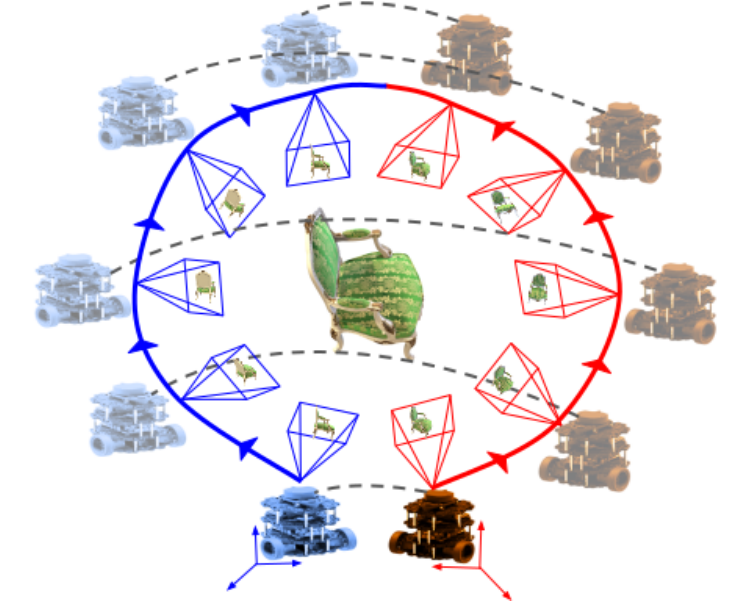}}
\vspace{-1 em}
\caption{Di-NeRF allows robots to cooperatively optimize local copies of a neural network model without explicitly sharing visual data. In this figure, two robots use Di-NeRF to cooperatively optimize a unified NeRF. Each robot only sees part of the chair, and robots do not know their relative poses. The robots communicate over a wireless network (gray dashed lines) to cooperatively optimize the final network and relative poses. 
}
\label{fig:dinerf_chair}
\end{figure}
There are many methods for distributed/centralized geometric multi-robot mapping and localization
~\cite{Murai2024TRO,Yulun2024TRO}. For learned methods, federated~\cite{9084352} and distributed learning~\cite{halsted2021survey} are common for overcoming the issues associated with centralized learning. In these methods, the training process is decentralized. Each robot performs its training, and the results are aggregated in a central node~\cite{9084352}. The need for a central node still constrains the scalability of such methods. A recent work using multiple agents for NeRF is Block-NeRF~\cite{tancik2022blocknerf}, composed of independent local maps, i.e. NeRF block. In Block-NeRF, the relative poses of the blocks are known to a degree and the task is done in a centralized manner.  
\textcolor{black}{For large-scale aerial NeRF mapping, Mega-NeRF~\cite{turki2022mega} partitions training images into NeRF submodules for parallel training.} 
Another closely related work is {DiNNO}~\cite{DBLP:journals/corr/abs-2109-08665}, \textcolor{black}{which 
is a distributed algorithm enabling robots to collaboratively optimize deep neural networks over a mesh network without sharing raw data.}
In this paper, we present Di-NeRF (Distributed NeRF), an algorithm that builds on the Consensus Alternating Direction Method of Multipliers (C-ADMM) as a versatile distributed optimization method for multi-robot systems. In Di-NeRF, each robot starts building its NeRF by relying on its own data \textcolor{black}{i.e. images and local camera poses from a Structure-from-Motion (SfM) algorithm}. 
By integrating NeRFs from other robots, each robot will build a global NeRF model as if the robot has gathered all data and processed them. The advantages of Di-NeRF are that the robots do not know their \textcolor{black}{exact} initial poses, 
{as illustrated graphically in Fig.~\ref{fig:dinerf_chair}.} 
In Di-NeRF, robots alternate between local optimization of an objective function and communication of intermediate network weights over the wireless network. Di-NeRF can consider different communication graphs (e.g. fully connected, circular, and ring connectivity). The key contributions of Di-NeRF include 
i)~developing fully distributed optimization for 3D reconstruction with RGB images as input, and a NeRF as the backend, enabling fusing NeRFs in training not the rendering process, and 
ii)~optimization of the relative poses of the robots, 
removing the need to know the prior relative poses \textcolor{black}{accurately}. \textcolor{black}{In this work, we make these assumptions: i) a reliable estimate of the agent's trajectory in its local frame is available; ii) there is sufficient overlap between the agents' viewpoints, allowing optimization of the relative poses.}

The structure of the paper is as follows. In Sec.~\ref{sec:lit_rev}, related work is presented. In Sec.~\ref{sec:method}, we present Di-NeRF, followed by results and conclusions in Sec.~\ref{sec:results} and~\ref{sec:conclusion}. 
\section{Literature Review} \label{sec:lit_rev}

Multi-robot localization and mapping algorithms utilize various representations such as sparse landmarks
~\cite{cunningham2013ddf}, dense geometric maps~\cite{schuster2019distributed}, object classes
~\cite{Tchuiev_2020}, and semantics
~\cite{tian2022kimera}. With the advent of neural radiance field~\cite{mildenhall2021nerf} and its variants 
\cite{wang2021nerfmm, yu2021pixelnerf, bian2023nope, mueller2022instant},
there are efforts to use neural maps for collaborative localization/mapping~\cite{tancik2022blocknerf, suzuki2023federated}.

Emerging radiance fields, including non-neural representations such as Plenoxels~\cite{yu_and_fridovichkeil2021plenoxels}, and neural representations, such as NeRF~\cite{mildenhall2021nerf, wu2023reconfusion} have revolutionized localization and mapping algorithms. Nice SLAM~\cite{Zhu2022CVPR}, iMAP~\cite{Sucar:etal:ICCV2021}, NeRF-SLAM~\cite{rosinol2023nerf}, and~\cite{teigen2024rgb} are using such representations. The early versions of NeRF representations~\cite{mueller2022instant}, 
required pose information often generated via Structure-from-Motion (SfM) algorithms, e.g. COLMAP~\cite{schoenberger2016sfm}. This limitation was later addressed in~\cite{wang2021nerfmm, bian2023nope, jeong2021selfcalibrating}, 
by jointly optimizing camera poses and the scene. It was also shown that a camera view can be localized in a NeRF map, as shown in iNeRF~\cite{yen2020inerf}.

Extending NeRF 
to multi-robot scenarios has also been proposed in Block-NeRF~\cite{tancik2022blocknerf}, NeRFuser~\cite{fang2023nerfuser}, and NeRFusion~\cite{zhang2022nerfusion}. These methods use blending techniques to render an image, often done centralized, performing inference on multiple NeRF models. Furthermore, it is assumed that there is prior knowledge about the relative pose of the robots. A similar research 
is federated learning~\cite{holden2023federated, suzuki2023federated}, 
where the training is done in a decentralized optimization, \textcolor{black}{and} each node learns a common NeRF in parallel. Then the weights are transferred to a server for aggregation. A memory-efficient multi-robot SLAM approach based on multi-block ADMM is proposed in~\cite{choudhary2015exactly}. \textcolor{black}{In DiNNO~\cite{DBLP:journals/corr/abs-2109-08665}, a distributed algorithm for collaborative learning of a neural model is presented, demonstrating a range of experiments including multi-agent RL, MNIST classification, and 2D neural implicit mapping with known relative poses.} The core of the algorithm is a distributed optimization method known as C-ADMM~\cite{halsted2021survey}. To establish a shared reference frame, \cite{indelman2016incremental} proposes a method for incremental inference from arbitrary poses. Additionally, im2nerf~\cite{mi2022im2nerf, yu2021pixelnerf} enables neural radiance field prediction from a single image, addressing view sparsity.

\section{Distributed Neural Radiance Field (Di-NeRF)} \label{sec:method}
\begin{figure}
\centering
\includegraphics[width=.9\linewidth]{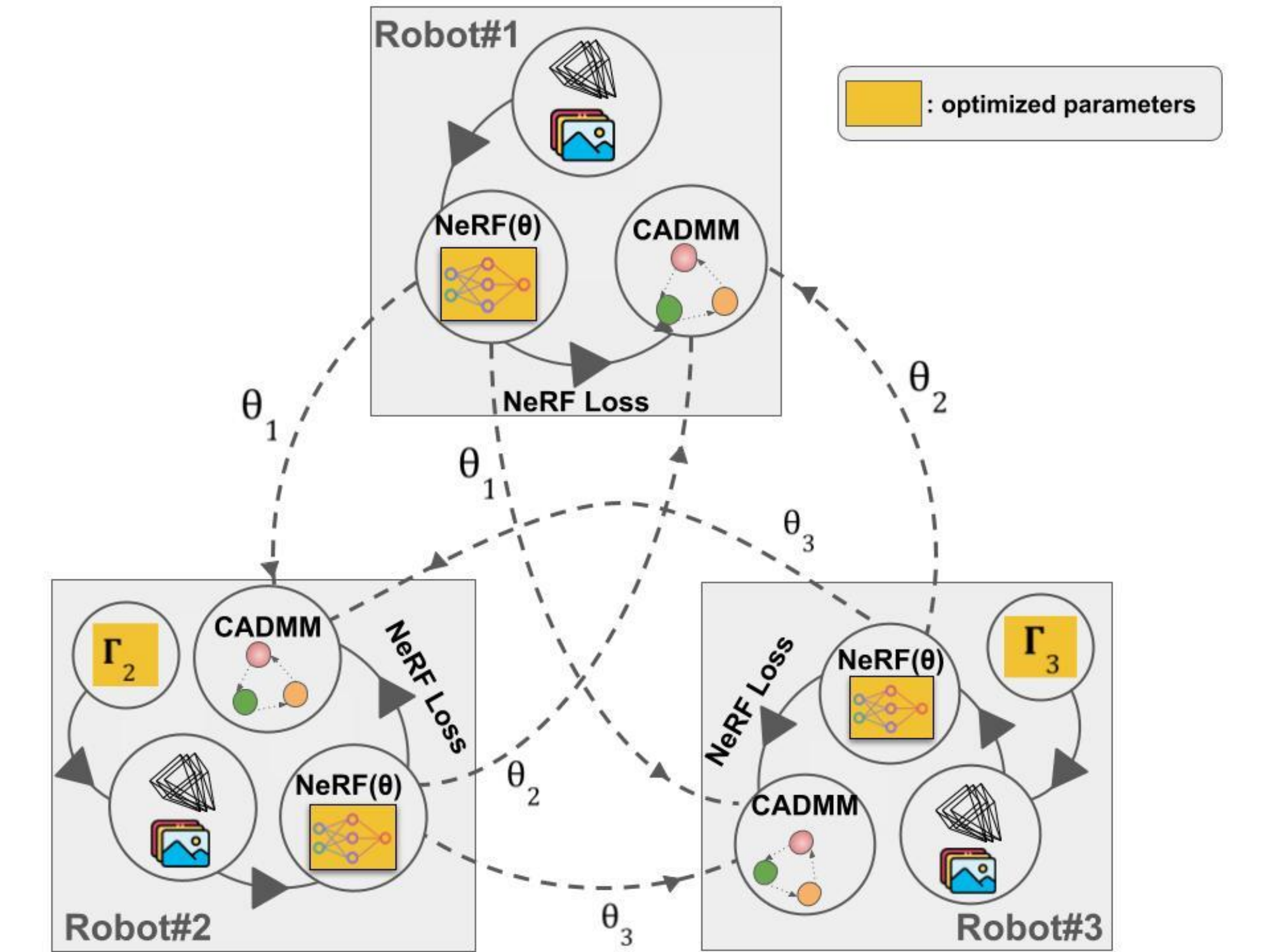}
\vspace{-2 mm}
\caption{
Three connected robots share their local NeRF to build a global NeRF. Each robot adjusts its weight compared to receiving weights using C-ADMM, and all robots except robot $R^g$ (Robot\#1, the robot whose coordinate is considered as the global coordinate) optimize for the relative pose $T_{i}^g$ and depth distortion parameters. 
}
\label{fig:dinerf_overview}
\end{figure}

Here the problem is defined followed by the proposed method for relative pose 
\textcolor{black}{refinement} and distributed NeRF.
The overall pipeline for Di-NeRF is shown in Fig.~\ref{fig:dinerf_overview}. 
Di-NeRF is a fully distributed algorithm for 3D reconstruction using RGB images. 
With Di-NeRF, each robot has its local coordinate and only shares the learned models 
to achieve a consensus while optimizing for the relative poses. 
\subsection{Problem Statement}
Assume $N$ robots connected to each other with a communication graph  $\mathcal{G}=(\mathcal{V},\mathcal{E} )$ consisting of vertices $\mathcal{V} = \{1,...,N\}$ and edges $\mathcal{E}$, on which pairwise communication can occur. \textcolor{black}{Given a set of images $\boldsymbol{I}_i$ captured \textcolor{black}{by robot $i$, $i=1,...,N$,} from a scene with the associated local camera poses, $\boldsymbol{\pi}_i$,} the goal of Di-NeRF is to build a scene representation, \textcolor{black}{$\theta$}, that enables all the robots to generate realistic images from novel viewpoints. 
\textcolor{black}{Here}, NeRF~\cite{mueller2022instant} is employed as the model for 3D reconstruction, \textcolor{black}{$\theta$}. 
Since the poses of each robot are defined in a local coordinate system, it is necessary to perform relative pose optimization to build a consistent model that accurately represents the entire scene. The problem is expressed as:
\begin{equation}\notag
        \textcolor{black}{\theta^*}, \textcolor{black}{\boldsymbol{T}^*} = \underset{\theta, \mathbf{T}}{\operatorname{argmin}}\sum_{i \in \mathcal{V}}{L_i(\theta, T^g_i \textcolor{black}{\mid}  \boldsymbol{I}_i, \boldsymbol{\pi}_i)},~
        \boldsymbol{T} = \{T^g_1,.., T^g_N\}, ~(1)
\label{problem_statement}
\end{equation}
\noindent where $L_i$ represents the NeRF loss function corresponding to the robot $i$. 
The relative poses are denoted by \(\boldsymbol{T}\), where the pose of the robot \(i\) relative to the global coordinate system is represented by \(T^g_i = [R^g|t^g]\) within the space of \(SE(3)\). Superscript $g$ stands for the global coordinate. \textcolor{black}{Without
loss of generality}, the first robot coordinate is selected as the global frame ($T^g_1$ is the identity matrix). \textcolor{black}{If this robot fails, another robot is selected as the global frame.}
\textcolor{black}{The model \(\theta\) is shared among all robots and needs to be determined collectively, indicated by the summation in Eq.~\eqref{problem_statement}. Other variables with subscript $i$; {\it i.e.} $T^g_i$, $\boldsymbol{I}_i$, and $\boldsymbol{\pi}_i$; are local/internal to robot $i$.} 
The model comprises fully connected neural network layers, described in Sec.~\ref{sec:details}. It learns to generate a color vector \(\boldsymbol{c} = (r, g, b)\) and volume density \(\sigma\) for each ray along a 3D location \(\boldsymbol{x} = (x, y, z)\) and 2D viewing direction $\boldsymbol{d}$. 
\textcolor{black}{Here, we use a pinhole camera model, assuming equal focal lengths for x and y directions.} 
The camera ray that starts from camera origin $\boldsymbol{o}$ and passes through each pixel $\boldsymbol{p}=(u,v)$ is in the local coordinate of each robot and should be transferred to the global coordinate system using relative camera poses. In a local camera pose, $\pi_i=[R|t]$, the camera origin is $\boldsymbol{o}=t$ and the ray direction $\boldsymbol{d}$ is as follows~\cite{mildenhall2021nerf, mueller2022instant}: 
\setcounter{equation}{1}
\begin{equation}
\boldsymbol{d} = R \begin{pmatrix}
\frac{u - W / 2}{f} &
\frac{\textcolor{black}{-}v \textcolor{black}{+} H / 2}{f} &
-1
\end{pmatrix}^{\top}~,
\end{equation}
\noindent where $f$ is the focal length and $H$ and $W$ are the image height and width. By using the relative pose $T^g_i=[R^g|t^g]$, the camera pose in the global coordinate will be $\pi_i^g = T^g_i \pi_i = [R^gR|R^gt+t^g]$ which positions the camera origin at $R^gt+t^g$ and defines ray direction $R^g\boldsymbol{d}$.
Since the relative poses, $\boldsymbol{T}^g_i$, are 
\textcolor{black}{not known exactly,} the model $\theta_i$ and relative poses must be optimized jointly. The optimization can be central, but a distributed approach is preferred for real-world use.

\subsection{Relative Camera Pose Optimization} \label{Relative_pose}
    
In multi-robot NeRF, robots do not send raw data over the wireless network. Therefore, using an SfM pipeline to calculate the relative poses for all robots is not easy. 
Here, the camera poses of each robot are expressed based on a local coordinate system, and gradually all poses transfer to the robot $R^g$ coordinate system (throughout this paper, the robot $R^g$ will be the arbitrary reference agent.
). The optimization of the relative poses is performed jointly with the model parameters. 
Relative camera poses of multi-robot systems can be expressed as a camera-to-world transformation matrix (similar to each robot's local camera pose)  \(T_i^g=[R^g|t^g]\) $\in$ \(SE(3)\), where \(R^g \in SO(3)\) and \(t^g \in \mathbb{R}^3\) show camera rotation and translation, respectively~\cite{wang2021nerfmm}. Optimizing the translation vector \(t^g\) involves the designation of trainable parameters, given its definition in Euclidean space. For camera rotation, which is in \(SO(3)\) space, the axis-angle representation is chosen: \(\phi:=\alpha \omega, \phi \in \mathbb{R}^3\), where \(\omega\) and \(\alpha\) are a normalized rotation axis and a rotation angle, respectively. The rotation matrix \(R^g\) can be expressed from Rodrigues’ formula: 
        \begin{equation}
        \label{Rodrigues’ formula}
        R^g = I + \frac{\sin{(\alpha)}}{\alpha}\phi^{\wedge} + \frac{1-\cos{(\alpha)}}{\alpha^2}(\phi^\wedge)^2 \ ,
    \end{equation}
where \((.)^\wedge\) is the skew operator that converts a vector \(\phi\) to a skew-symmetric matrix. The relative pose for each robot $i$ is optimized by trainable parameters \(\phi_i\) and \(t_i\).

\textcolor{black}{Training NeRF and the relative camera poses jointly adds another degree of ambiguity to the problem, and causes the NeRF optimization to converge to local minima, especially for large relative poses. 
To avoid local minima, we use monocular depth estimation~\cite{bian2023nope}. However, the depth maps suffer scale and shift distortions across frames, making mono-depth maps not multi-view consistent. 
To utilize mono-depth effectively, we explicitly optimize scale and shift parameters during NeRF training by penalizing the difference between rendered depth and mono-depth. 
For each image $I^m_i$ (out of $M_i$ images) of robot $i$, 
we generate mono-depth sequence $D^m_i, m= 0, ..., M_{i}-1$ from input images 
with an off-the-shelf monocular depth network, i.e., DPT~\cite{ranftl2021vision}. Two linear transformation parameters $\boldsymbol{\psi}_i^m=(\alpha^m_i,\beta^m_i), m=0 \ldots M_{i}-1$ for each mono-depth map as scale and shift factors can recover a multi-view consistent depth map ${D^m_i}^* = \alpha^m_i D^m_i + \beta^m_i$, for each image $m$ in each robot $i$. This joint optimization of distortion parameters and NeRF is achieved by adding a depth loss $L^{depth}_i = \sum_{m}||{D^m_i}^*-\hat{D}^m_i||$ to the RGB loss of NeRF~\cite{mueller2022instant}, $L^{RGB}_i = \sum_{m}||{I^m_i}-\hat{I}^m_i||$. Here $\hat{D}^m_i$ and $\hat{I}^m_i$ are NeRF-rendered depth/RGB images, \textcolor{black}{described in~\cite{bian2023nope},~\cite{mueller2022instant}}. Thus, the local NeRF loss for each robot $i$ is defined as:
    \begin{equation}
    \label{depth_loss}
        L_i = L^{RGB}_i + \gamma L^{depth}_i~,
    \end{equation}
\noindent where $\gamma$ is a weighting factor for depth loss term. With depth information, Eq.~\eqref{problem_statement} is rewritten as:
\begin{equation}
        \theta^*, \boldsymbol{T}^*, \boldsymbol{\psi}^* = \underset{\theta, \mathbf{T}, \boldsymbol{\psi}}{\operatorname{argmin}}\sum_{i \in \mathcal{V}}{L_i(\theta, T^g_i, \boldsymbol{\psi}_i \mid  \boldsymbol{I}_i, \boldsymbol{\pi}_i)}~.
\label{problem_statement2}
\end{equation}
$\boldsymbol{\psi_i} = \{{\psi}_i^m\}$ is the set of distortion parameters for robot $i$, and $\boldsymbol{\psi} = \{\boldsymbol{\psi}_i\}$ is the set of all distortion parameters of all robots. }
\subsection{Distributed Formulation}
In the problem presented in Eq.~\eqref{problem_statement2}, each agent $i$ has its local cost function $L_i$, indicating that the optimization must be done in a distributed manner on each agent. However, $\theta$ is a global variable and is shared among all agents. This necessitates that the agents should achieve a consensus on the optimal value of $\theta$. In addition to the global variable $\theta$, agents have their local variables, such as $T_i^g$. To express that the optimization is distributed, 
Eq.~\eqref{problem_statement2} is rewritten as 

\begin{equation}
\label{problem_new}
\begin{aligned}
\textcolor{black}{\theta^*, \boldsymbol{T}^*, \boldsymbol{\psi}^*}&= \underset{\theta, \boldsymbol{T}, \textcolor{black}{\boldsymbol{\psi}}}{\operatorname{argmin}} \sum_{i \in \mathcal{V}} L_i(\underbrace{\theta_i, T_i^g, \textcolor{black}{\boldsymbol{\psi}_i}}_{{\textcolor{black}{\Gamma_i}}} \textcolor{black}{\mid} \boldsymbol{I}_i, \boldsymbol{\pi}_i )~. \\
\text{s.t.} \quad \theta_i &= z_{ij}, \quad \forall j \in \mathcal{N}_i
\end{aligned}
\end{equation}

\noindent Each agent solves for its version of the global variable $\theta$, $\theta_i$, a relaxed local variable or primal variable. Assuming $\mathcal{N}_i$ is the set of neighbors of robot $i$, to achieve a consensus between agents $i$ and $\forall j \in \mathcal{N}_i$, an auxiliary variable $z_{ij}$ is introduced. This is also known as the `complicating variable’ which ensures that the agents achieve a consensus via `complicating constraints’: i.e. $\theta_i = z_{ij}$ and $\theta_j = z_{ij}$. 
\textcolor{black}{By defining ${\Gamma}_i = \{\theta_i, T_i^g, \boldsymbol{\psi}_i\}$ for notation brevity}, 
Eq.~\eqref{problem_new} 
is then solved by introducing augmented Lagrangian and redefining the cost function to $\mathcal{L}_i$ \textcolor{black}{-- for brevity, $\boldsymbol{I}_i$ and $\boldsymbol{\pi}_i$ were omitted}: 
\begin{equation}
\begin{aligned}
\mathcal{L}_i(\textcolor{black}{{\Gamma}_i}) = L_i(\textcolor{black}{{\Gamma}_i})+( \theta_i-z_{ij})^{\top} y_i + \frac{\rho}{2} \sum_{j \in \mathcal{N}_i }{\left\| \theta_i-z_{ij}\right\|_2^2} \ ,
\end{aligned}
\label{eq:augmented_lagrangian}
\end{equation}
\noindent where $y_i$ is the Lagrangian multiplier or dual variable and $\rho$ is penalty factor. The last two terms enforce that constraints are satisfied, but the penalty function ensures that around the optimal point, the objective function is quadratic. 

There are various techniques to optimize Eq.~\eqref{eq:augmented_lagrangian}, such as auxiliary problem principle (APP)~\cite{losi_2003_app} and alternating direction method of multipliers (ADMM)~\cite{Boyd2011}. ADMM has improved convergence properties~\cite{Boyd2011}. We use a version of ADMM that is suitable for distributed systems, known as consensus ADMM (C-ADMM)~\cite{shorinwa2023distributedpt2} described next.

\subsection{Distributed Optimization of NeRF}

The optimization through ADMM alternates between variables, and C-ADMM allows the agents to share the intermediate optimized variables to achieve a consensus. The variables are \textcolor{black}{${\Gamma}_i = \{\theta_i, T_i^g, \boldsymbol{\psi}_i\}$}, $z_{ij}$, and $y_i$. \textcolor{black}{Optimization of $z_{ij}$ has a closed-form solution, 
}
see the website of the project, 
(superscript (k) denotes the step number in updating the variable):
\begin{equation}
z_{ij}^{(k+1)} = \frac{1}{N}\sum\theta_i^{(k)} \coloneqq \bar{\theta}^{(k)}~,
\end{equation}
\textcolor{black}{therefore,} 
the optimization step alternates between minimizing the updated local objective function with respect to primal variables (\textcolor{black}{
${\Gamma}_i = \{\theta_i, T_i^g, \boldsymbol{\psi}_i\}$}) and maximizing the updated local objective function with respect to the dual variable:

$\bullet~\theta_i$-minimization 
\begin{equation}
\underset{\textcolor{black}{{\Gamma}_i^{(k)}}}{\operatorname{min}}~L_i(\textcolor{black}{{\Gamma}_i^{(k)}}) + \theta_i^{(k)\top} y_i^{(k)}  + \frac{\rho}{2} \sum_{j \in \mathcal{N}_i}\left\|\theta^{(k)}_i- \bar{\theta}^{(k)}\right\|_2^2,
\label{eq-C-ADMM}
\end{equation}

$\bullet~y_i$-update (or dual variable update):
\begin{equation}
y_i^{(k)} \gets y_i^{(k-1)} + \rho \sum_{j \in \mathcal{N}_i}{(\theta_i^{(k)}-\bar{\theta}^{(k)})},
\end{equation}

Alg.~\ref{dinerf} summarizes the optimization procedures for local primal, dual variables, and relative poses for each robot. 
In line 1, The initial values for the NeRF parameters $\theta_i^{(0)}$, the dual variable $y_i^{(0)}$, \textcolor{black}{and the depth distortion parameters $\boldsymbol{\psi}_i^{(0)}$} are set to zero. 
The initial value for the relative poses - three rotation angles $R$ and three translation displacements $t$ - are explained in Sec.~\ref{sec:results}. 
\textcolor{black}{The first robot $R^g$ undergoes the NeRF training process (ln 6-7), communicates the learned weights with other robots (ln 10-11), and updates the dual variable (ln 12) for 200 steps}. The remaining robots engage in collaborative optimization, simultaneously refining NeRF weights and adjusting relative camera poses \textcolor{black}{and depth distortion parameters} 
in a distributed training framework based on the weights of robot $R^g$ \textcolor{black}{(lines 8-9 for NeRF training, lines 10-12 for communication and dual update)}. This iterative communication and training process persists until uniformity is achieved in both the weights and optimized relative camera poses across all robots. The relative pose between robot \(i\)'s coordinate system and robot $R^g$'s coordinate system 
\(T_i^g \in SE(3)\) is optimized jointly in the distributed training \textcolor{black}{(ln 9)}.

\begin{algorithm}
    \caption{Di-NeRF Algorithm}
    \label{dinerf}
    \begin{algorithmic}[1] 
        \State \textbf{Initialization:} $k \leftarrow 0, \theta_i^{(0)} \in \mathbb{R}^n, y_i^{(0)}=0, T_i^g \in SE(3)$, \textcolor{black}{$\boldsymbol{\psi_i} \in \mathbb{R}^m$}
        \State \textbf{Internal variables:} $y_i^{(k)}$  
        \State \textbf{Public variables:} $\mathcal{Q}_i^{(k)}=\theta_i^{(k)}$  \label{line:communicated}
        
        \While{stopping condition not satisfied}
            \For{$i$ in $\mathcal{V}$}  
                \If{robot $i$ is $R^g$}
                    \State $\theta_i^{(k+1)} = \underset{\theta_i}{\operatorname{argmin}}\{L_i\left(\theta_i\right) + \theta_i^{\top} y_i^{(k)}$  
                    \Statex  \ \ \ \ \ \ \ \ \ \ \ $+\rho \sum_{j \in \mathcal{N}_i}\left\|\theta_i-\bar{\theta}^{(k)}\right\|_2^2\}$  
                \Else
                    \State $\textcolor{black}{{\Gamma}}_i^{(k+1)} = \underset{\textcolor{black}{{\Gamma}}}{\operatorname{argmin}}$   
                    \Statex \ \ \ \ \ \ \ \ \ $\{L_i\left(\textcolor{black}{{\Gamma}}_i\right)+\theta_i^{\top} y_i^{(k)}+\rho \sum_{j \in \mathcal{N}_i}\left\|\theta_i-\bar{\theta}^{(k)}\right\|_2^2\}$  
                \EndIf
                \State \textbf{Communicate} $\mathcal{Q}_i^{(k)}$ to all $j$ in $\mathcal{N}_i$
                \State \textbf{Receive} $\mathcal{Q}_j^{(k)}$ from all $j$ in $\mathcal{N}_i$
                \State $y_i^{(k+1)} = y_i^{(k)} + \rho \sum_{j \in \mathcal{N}_i}(\theta_i^{(k+1)} - \theta_j^{(k+1)})$  
                \State $k \leftarrow k+1$  
            \EndFor  
        \EndWhile  
    \end{algorithmic}
\end{algorithm}

\subsection{Convergence Properties}
The convergence of C-ADMM methods  
typically requires the dual variables' sum to converge to zero, a condition challenging in unreliable networks. Moreover, the nonlinearity and nonconvexity of neural networks, especially NeRF, preclude guaranteed global solutions and linear rates~\cite{DBLP:journals/corr/abs-2109-08665, shorinwa2023distributedpt2, fridovich2023gradient}. Despite these issues, Alg.~\ref{dinerf} has proven effective in practice for distributed NeRF training, as shown in Sec.~\ref{sec:results}, converging to solutions equivalent to when a robot is doing all processing.

\section{Experiments} \label{sec:results}
This section presents the experimental results on synthetic and real-world datasets, including  
1)~comparing Di-NeRF and a baseline, 
2)~examining the impact of the number of robots and communication graphs, 
3)~analyzing the impact of the amount of overlap between the robots' trajectories on convergence,  
4)~evaluation using Waymo Block-NeRF dataset~\cite{tancik2022blocknerf}, 
\textcolor{black}{5)~convergence of the relative pose with respect to different initial values and ground truth relative pose,}
\textcolor{black}{6)~examining no overlap setup,} 
\textcolor{black}{7)~assessing reference robot failure, and 
8) an ablation study.} 
The architecture of NeRF is based on the instant neural graphics primitives (iNGP)~\cite{mueller2022instant}, described in Sec~\ref{sec:details}.
%
For all comparisons, the `\emph{baseline}' is training one model with all images, with known poses, equivalent to the standard NeRF training of iNGP. 
For the relative poses, the initial values for both the translational and rotational components were \textcolor{black}{set to zero, except for convergence analyses (Sec.~\ref{sec:E} and Sec.~\ref{sec:abalation}, Fig.~\ref{fig:abalation}). The ground truth relative poses assume each robot's local origin is its first frame, chosen for the largest relative pose. In reality, any image in the robot's trajectory could serve as the origin.} 
\subsection{Di-NeRF vs \textcolor{black}{Baseline} NeRF on Synthetic Dataset}
\label{sec:A}
In this experiment, we use Di-NeRF to learn a model in a distributed manner and compare the results with the baseline, iNGP~\cite{mueller2022instant}. 
\textcolor{black}{In this experiment, for Di-NeRF,} the robots do not have access to a global coordinate frame, so the relative poses are unknown.
\textcolor{black}{We also compare Di-NeRF with PixelNeRF~\cite{yu2021pixelnerf}. PixelNeRF works with one/fewer images. \textcolor{black}{So part of the data from both robots that produce the best results is considered for PixelNeRF.}}
%
Here, the Chair, Hotdog, and Lego sequences from the synthetic NeRF dataset~\cite{mildenhall2021nerf} are used. We also use the long sequence of Barn from the real-world Tanks and Temples Benchmark~\cite{Knapitsch2017}.
The configurations for all experiments and the \textcolor{black}{baseline} are the same, described in Sec.~\ref{sec:details}. 
The \textcolor{black}{baseline} contains 100-150 images for synthetic sequences and 300 for the Barn sequence. For multiple robots, images are divided with controlled overlaps. All experiments follow official pre-processing and train/test splits. 

\begin{figure}[t!]
    \vspace{-1.2em}
    \centering
    \color{black}
    \includegraphics[width=1\linewidth]{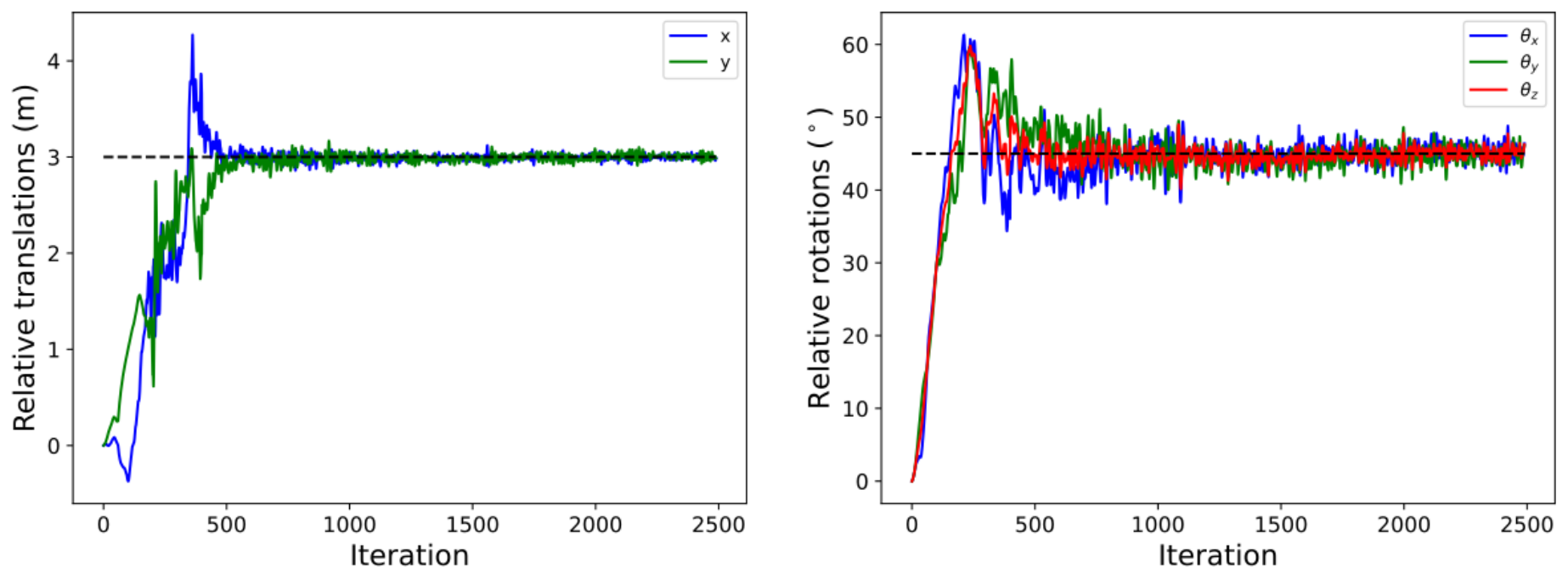}
    \vspace{-2em}
    \caption{\textcolor{black}{Optimizing the relative poses for two robots for the chair sequence~\cite{mildenhall2021nerf}. The relative translation and rotation estimates are plotted. The desired translation/rotation is marked by a dashed line.}}
    \label{fig:nerf_rel_pose_chair}
\end{figure}

The metrics used to compare the performance of Di-NeRF with the \textcolor{black}{baseline} solution are the average rendering i)~PSNRs, ii)~SSIM, and iii)~\textcolor{black}{relative} 
pose accuracy (in degrees and cm). The pose accuracy is done against the ground truth. 
For the synthetic sequences, two robots split the dataset, with robot $R^1$ observing the chair's front and robot $R^2$ the back, without trajectory overlap. This pattern applies to other sequences. The dataset is divided, and COLMAP estimates poses for each segment, with different coordinate systems. Di-NeRF calculates the relative pose. Fig.~\ref{fig:nerf_rel_pose_chair} shows the convergence of the relative pose for the chair sequence.
\begin{figure*}[t!]
    \vspace{1em}
    \centering
    \color{black}
    \includegraphics[width=1\linewidth]{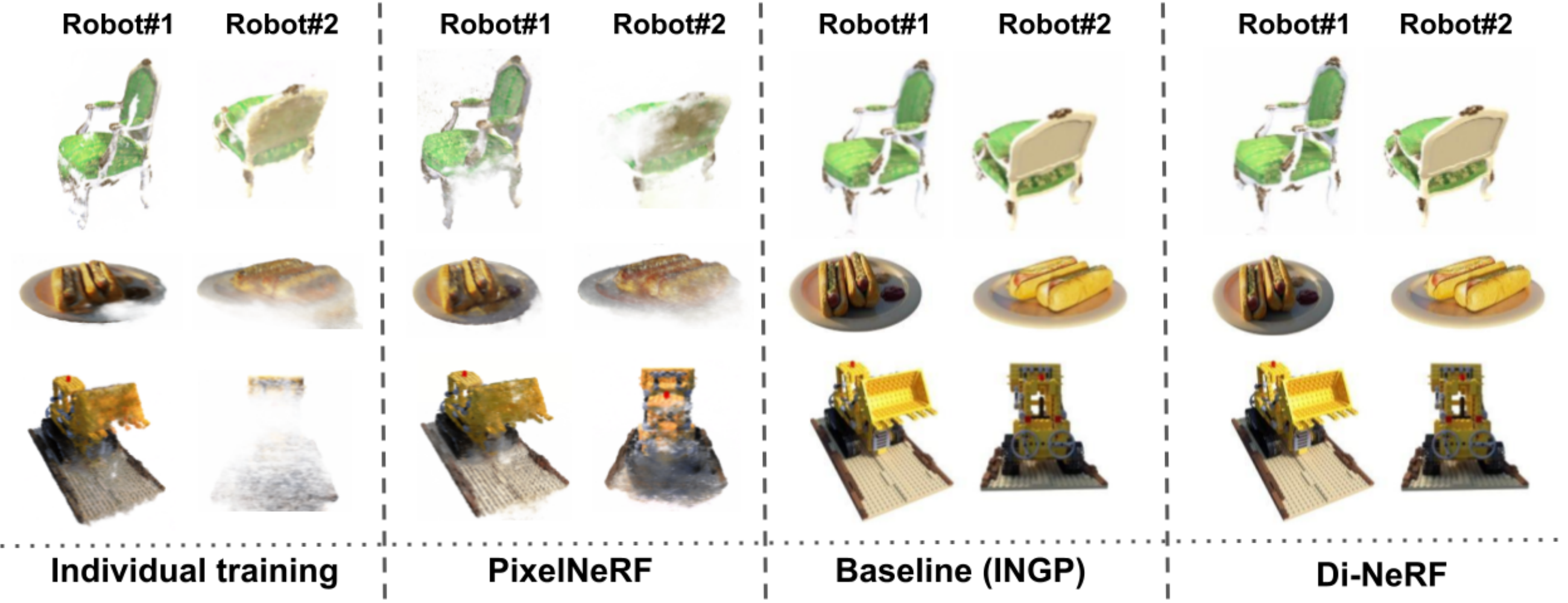}
    \vspace{-2em}
    \caption{\textcolor{black}{Two robots learn the scene collaboratively 
             where each robot only sees a part of the scene. Individual training results \textcolor{black}{(using iNGP and PixelNeRF)} have poor quality for some areas, 
             but Di-NeRF maintains good quality, similar to the baseline. In each column, the left and right images are for robots $R^1$ and $R^2$, \textcolor{black}{respectively.}}}
    \label{fig:nerf_chair}
\end{figure*}

\begin{table*}[t!]
    \centering
    \caption{{Image and relative camera pose metric performances for two robots. 
    PSNR and SSIM are reported for each robot. The relative pose accuracy is shown for $R^2$. $R^1$ is the global frame. {lower/higher better is indicated by $\downarrow$/$\uparrow$ respectively.}}}
    \vspace{-1 em}
    \label{table1}
    \begin{tabular}{|c|c|c|c|c|c|c|c|c|c|c|c|c|c|}
    \hline
    \small
    \multirow{2}{*}{Dataset}& \multirow{2}{*}{$T_i^g$} &\multicolumn{4}{|c|}{Di-NeRF} &\multicolumn{4}{|c|}{Centralized} & \multicolumn{2}{|c|}{Baseline}& \multicolumn{2}{|c|}{\textcolor{black}{PixelNeRF}} \\
    \cline{3-14}
     & & PSNR$\uparrow$ & SSIM$\uparrow$ & $\delta R^{\circ}\downarrow$ & $\delta t(cm)\downarrow$  & PSNR& SSIM & $\delta$ $R^{\circ}$& $\delta t (cm)$ & PSNR & SSIM &\textcolor{black}{PSNR} & \textcolor{black}{SSIM}\\
    \hline
    \multirow{2}{*}{Chair}&Known&31.93/32.05&0.941/0.945&NA&NA&33.02&0.952&NA&NA&\multirow{2}{*}{33.02}&\multirow{2}{*}{0.952}&\multirow{2}{*}{\textcolor{black}{24.61}}&\multirow{2}{*}{\textcolor{black}{0.682}}\\
    \cline{2-10}
    &Unknown&\textcolor{black}{32.11/31.89}&\textcolor{black}{0.933/0.939}&\textcolor{black}{1.2}&\textcolor{black}{2.02}&32.63&0.923&0.11&0.86&&&&\\
    \hline
    \multirow{2}{*}{Hotdog}&Known&31.02/31.10&0.951/0.952&NA&NA&32.63&0.963&NA&NA&\multirow{2}{*}{32.63}&\multirow{2}{*}{0.963}&\multirow{2}{*}{\textcolor{black}{24.11}}&\multirow{2}{*}{\textcolor{black}{0.695}}\\
    \cline{2-10}
    &Unknown&\textcolor{black}{30.25/30.50}&\textcolor{black}{0.931/0.917}&\textcolor{black}{1.30}&\textcolor{black}{1.78}& 32.21 &0.958& 0.54& 0.81&&&&\\
    \hline
    \multirow{2}{*}{Lego}&Known&29.00/28.91&0.936/0.937&NA&NA&29.67&0.941&NA&NA&\multirow{2}{*}{29.67}&\multirow{2}{*}{0.941}&\multirow{2}{*}{\textcolor{black}{25.07}}&\multirow{2}{*}{\textcolor{black}{0.706}}\\
    \cline{2-10}
    &Unknown&\textcolor{black}{29.11/29.19}&\textcolor{black}{0.932/0.938}&\textcolor{black}{1.42}&\textcolor{black}{1.22}&29.71 &0.943& 0.51& 0.76&&&&\\
    \hline
    \multirow{2}{*}{Barn}&Known&29.01/28.79&0.884/0.886&NA&NA&29.94&0.894&NA&NA&\multirow{2}{*}{29.94}&\multirow{2}{*}{0.894}&\multirow{2}{*}{\textcolor{black}{24.16}}&\multirow{2}{*}{\textcolor{black}{0.659}}\\
    \cline{2-10}
    &Unknown&\textcolor{black}{29.09/29.23}&\textcolor{black}{0.854/0.843}&\textcolor{black}{1.42}&\textcolor{black}{1.70}&29.66&0.891 &0.59 & 0.94&&&&\\
    \hline
    \end{tabular}
\end{table*}
\begin{figure}[t!]
    \centering
    \color{black}
    \includegraphics[width=1\linewidth]{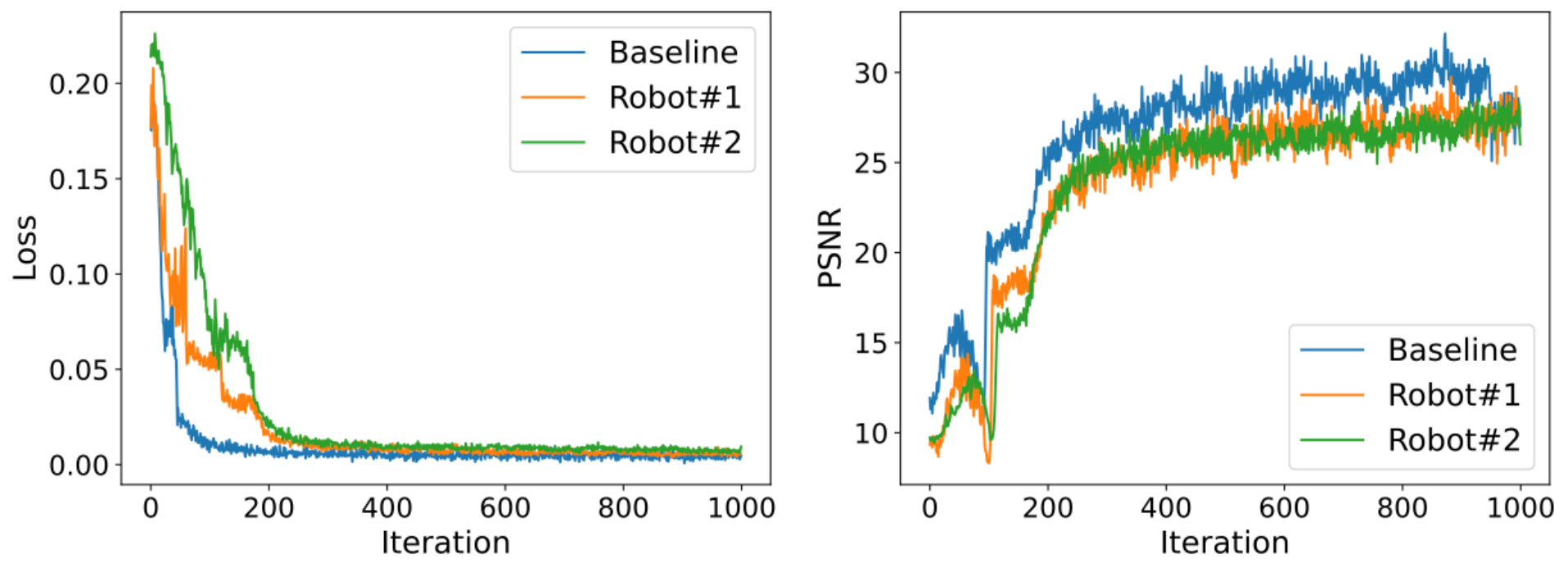}
    \vspace{-2 em}
    \caption{\textcolor{black}{NeRF validation Loss and PSNR for the chair sequence. 
    }}
    \label{fig:nerf_val_loss}
\end{figure}
Fig.~\ref{fig:nerf_chair} shows the qualitative results. 
\textcolor{black}{It illustrates the Chair sequence, comparing individual training, PixelNeRF, the baseline, and Di-NeRF for two robots. Since robot $R^1$ has no image from the back of the chair, it fails to reconstruct the back. Similarly, robot $R^2$ is unable to accurately reconstruct the front of the chair. This issue is evident in the first column of Fig.~\ref{fig:nerf_chair}.}
The \textcolor{black}{second} column is PixelNeRF. The \textcolor{black}{ third} column shows the baseline. Di-NeRF enables all robots to render the whole scene, see Fig.~\ref{fig:nerf_chair}-(last column). 
Fig.~\ref{fig:nerf_val_loss} presents a comparison of PSNR and L2 loss convergences for Di-NeRF and the \textcolor{black}{baseline} training. The convergences exhibit nearly identical behavior. 

Table~\ref{table1} shows the quantitative results for PixelNeRF, a centralized method, the baseline, and Di-NeRF. The metrics separated by the / sign show the metric values for each robot. 
The \emph{centralized} column refers to sending all local images and local poses to a server for training a NeRF model. The relative poses are calculated from RootSIFT features of a set of keyframes from each local set. \textcolor{black}{The global poses then are calculated to train a model.}
Di-NeRF achieves similar quality compared to the baseline and centralized methods. 
\textcolor{black}{For iNGP and PixelNeRF, image poses are in the global frame. Table~\ref{table1} shows that while PixelNeRF is designed for a few input images, Di-NeRF outperforms it with results matching centralized and baseline approaches.}

\begin{figure}[t!]
    \vspace{-1.2 em}
     \centering
     \color{black}
     \includegraphics[width=1\linewidth]{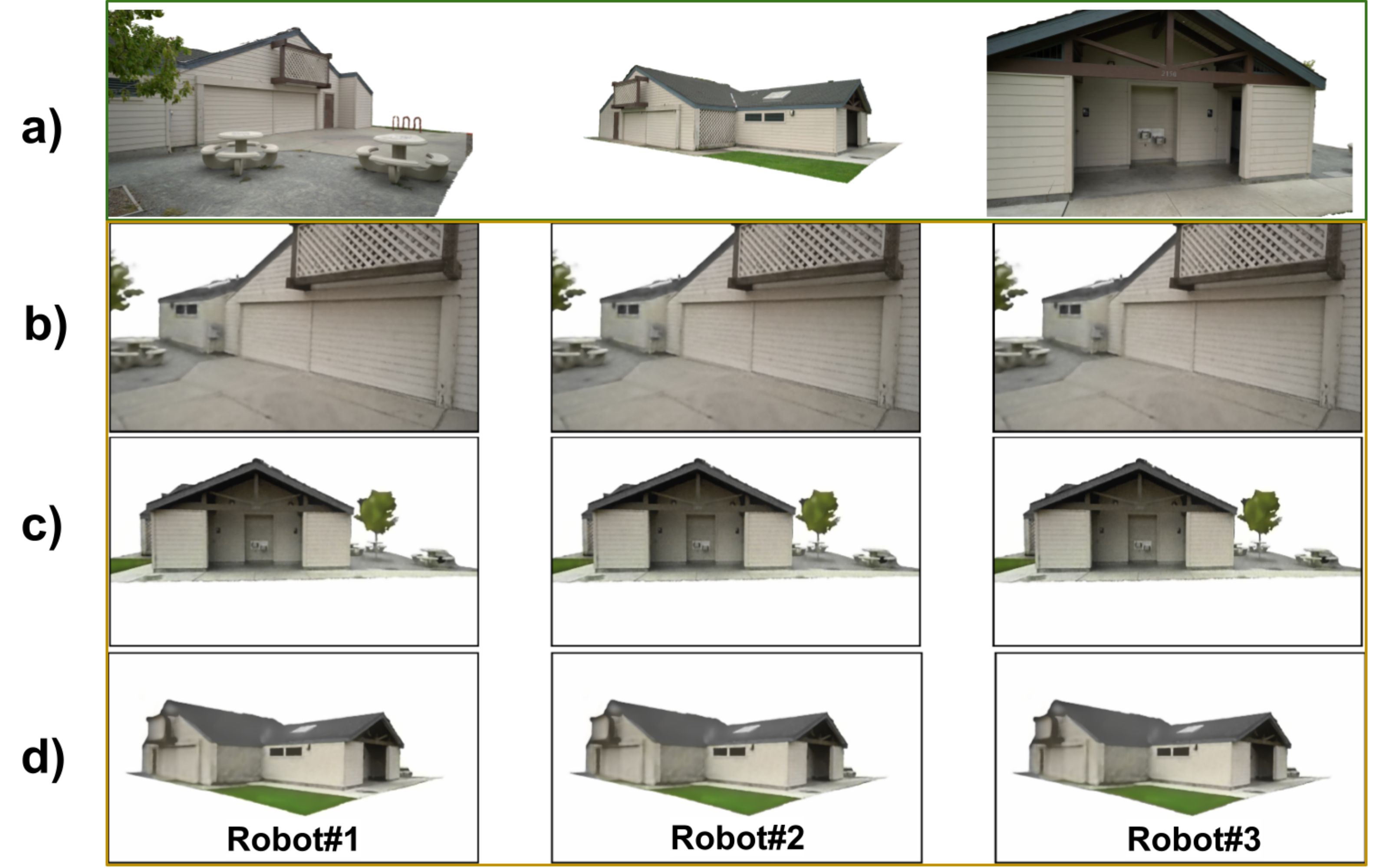}
     \vspace{-2 em}
     \caption{\textcolor{black}{(a) sample images, from the Barn sequence, provided to each robot; (b, c, d) sample images generated by Di-NeRF, once the robots collaboratively learn the scene. The images are strategically chosen—each is only visible in the raw data of a specific robot but can be rendered by all robots.}}
     \label{fig:barn_renders}
     \vspace{1 em}
     \color{black}
     \centering
     \includegraphics[width=1\linewidth]{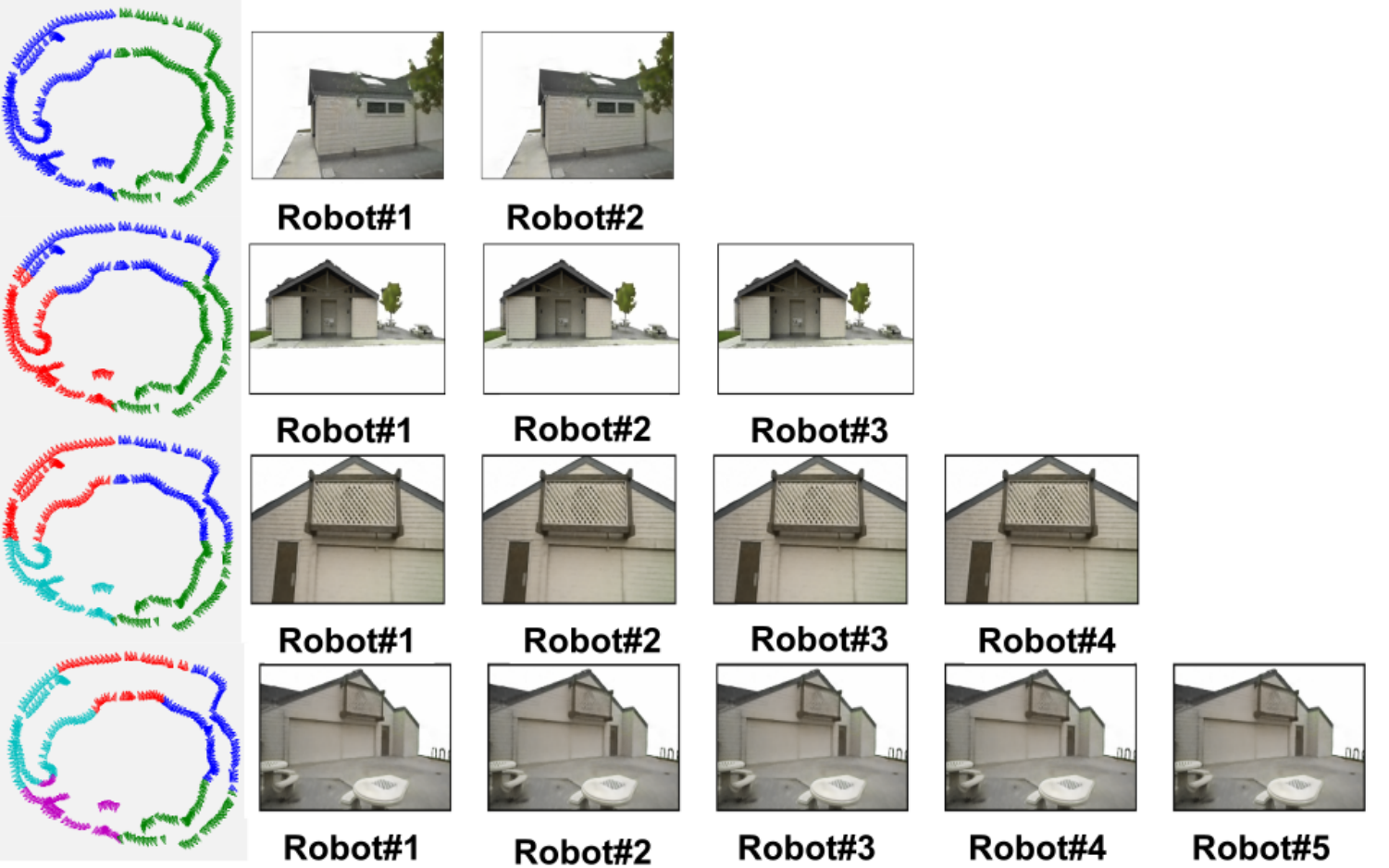}
     \vspace{-2 em}
     \caption{\textcolor{black}{Di-NeRF for different numbers of robots (all fully connected); 
     (left) the allocation of frames and poses 
     to the robots; 
     (right) none of the robots have common images, but 
     all robots can render the whole scene.}}
     \label{fig:barn_robots}
      \vspace{-1.3 em}
\end{figure} 
\subsection{Di-NeRF vs \textcolor{black}{Baseline} NeRF on Real-world Dataset}
\label{sec:B}
This experiment shows the performance of Di-NeRF not only on real-world data, but also when the number of robots increases from two to five. This scenario applies to large-scale environments. 
Here, the Barn sequence~\cite{Knapitsch2017} is used. There are 300 images in this sequence. Images are divided by the number of robots such that there is no common image in the local data of each robot, and the robots' trajectories have zero overlaps. Fig.~\ref{fig:barn_renders} shows the rendered images from each robot's distributedly trained NeRF. Each robot can render a 
view from any point 
\textcolor{black}{ (e.g.} Robot $R^3$ has never seen the large bay doors but renders an image when requested, as seen in Fig.~\ref{fig:barn_renders} (b)).
    
Next, the number of robots in this experiment is altered. In Fig.~\ref{fig:barn_robots}, the trajectory for each robot is shown alongside the rendered images for each robot after collaborative training is completed. Each color represents the trajectory of a different robot. Table~\ref{table2} presents the quantitative results. The rendering qualities and accuracy of relative camera pose are compared to the \textcolor{black}{baseline} method. With Di-NeRF, the outcomes closely resemble those achieved with the \textcolor{black}{baseline} method. In Table~\ref{table3}, we experiment with different connectivity graphs of robots. Five robots are used with the Barn sequence. In the star connection, $R^1$ is in the center. In Table~\ref{table3}, the average results for 5 robots are shown.

\begin{table*}
    \vspace{1 em}
    \centering
    \caption{Image and relative camera pose metric performances for different numbers of robots for the Barn sequence. 
    }
    \vspace{-1 em}
    \label{table2}
    \begin{tabular}{|c|c|c|c|c|c|c|}
    \hline
    & Di-NeRF & Baseline & Di-NeRF & Baseline & \multicolumn{2}{|c|}{Relative Camera Pose Error} \\
    \hline
    \#Robots& \multicolumn{2}{|c|}{PSNR $\uparrow$} & \multicolumn{2}{|c|}{SSIM $\uparrow$} & $\delta R^{\circ}$ $\downarrow$ & $\delta t (cm)$ $\downarrow$ \\ 
    \hline
    2& \textcolor{black}{29.21/29.11} & \multirow{4}{*}{29.94} & \textcolor{black}{0.886/0.839} & \multirow{4}{*}{0.894} & \textcolor{black}{1.34} & \textcolor{black}{1.76} \\
    \cline{1-2} \cline{4-4} \cline{6-7}
    3& \textcolor{black}{29.13/29.21/29.08} &  & \textcolor{black}{0.836/0.878/0.873} &  & \textcolor{black}{2.12/1.89} & \textcolor{black}{1.05/1.82} \\
    \cline{1-2} \cline{4-4} \cline{6-7}
    4& \textcolor{black}{29.35/29.39/29.33/29.17} & & \textcolor{black}{0.856/0.838/0.828/0.856} &  & \textcolor{black}{1.35/1.90/2.10} & \textcolor{black}{1.83/2.00/3.01} \\
    \cline{1-2} \cline{4-4} \cline{6-7}
    5& \textcolor{black}{29.67/29.44/29.15/29.45/29.41} &  & \textcolor{black}{0.845/0.878/0.852/0.887/0.869} &  & \textcolor{black}{2.70/3.20/3.10/3.00} & \textcolor{black}{2.11/2.65/2.39/2.02} \\
    \hline
    \end{tabular}
        \vspace{-1.7 em}
\end{table*}

\begin{table}
\vspace{-0.5 em}
    \centering
\caption{Mean validation result of Di-NeRF for 5 robots, with different connectivities, for the Barn sequence. Di-NeRF works well even if the network of robots is not fully connected. \textcolor
{black}{lower/higher better is indicated by $\downarrow$/$\uparrow$ respectively.}}
\vspace{-1 em}
\label{table3}
    \begin{tabular}{|c|c|c|c|c|} \hline 
          Communication&  PSNR$\uparrow$ & SSIM$\uparrow$ & $\delta R^\circ$ $\downarrow$ & $\delta t (cm)$ $\downarrow$ \\ \hline 
         Fully Connected&  \textcolor{black}{29.42}&  \textcolor{black}{0.866}&  \textcolor{black}{3.00}&\textcolor{black}{2.29}\\ \hline 
         Ring&  \textcolor{black}{29.19}&  \textcolor{black}{0.835}&  \textcolor{black}{4.56}&\textcolor{black}{2.33}\\ \hline 
         Star&  \textcolor{black}{29.16}&  \textcolor{black}{0.862}&  \textcolor{black}{4.91}&\textcolor{black}{3.02}\\ \hline 
         Line&  \textcolor{black}{29.10}&  \textcolor{black}{0.853}&  \textcolor{black}{5.01}&\textcolor{black}{2.97}\\ \hline
    \end{tabular}
    \vspace{-1 em}
\end{table}

\begin{figure}[t!]
     \vspace{-1 em}
    \centering
    \color{black}
    \includegraphics[width=1\linewidth]{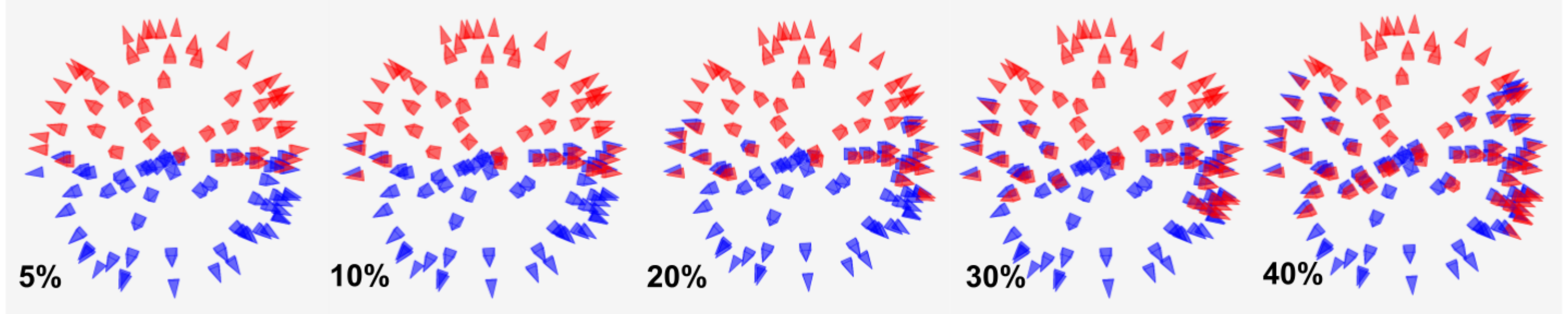}
    \vspace{-2 em}
    \caption{Segmenting the Chair sequence with different overlaps. 
    }
    \label{fig:chair_overlaps}
       \vspace{1 em}
    \centering
    \color{black}
    \includegraphics[width=1\linewidth]{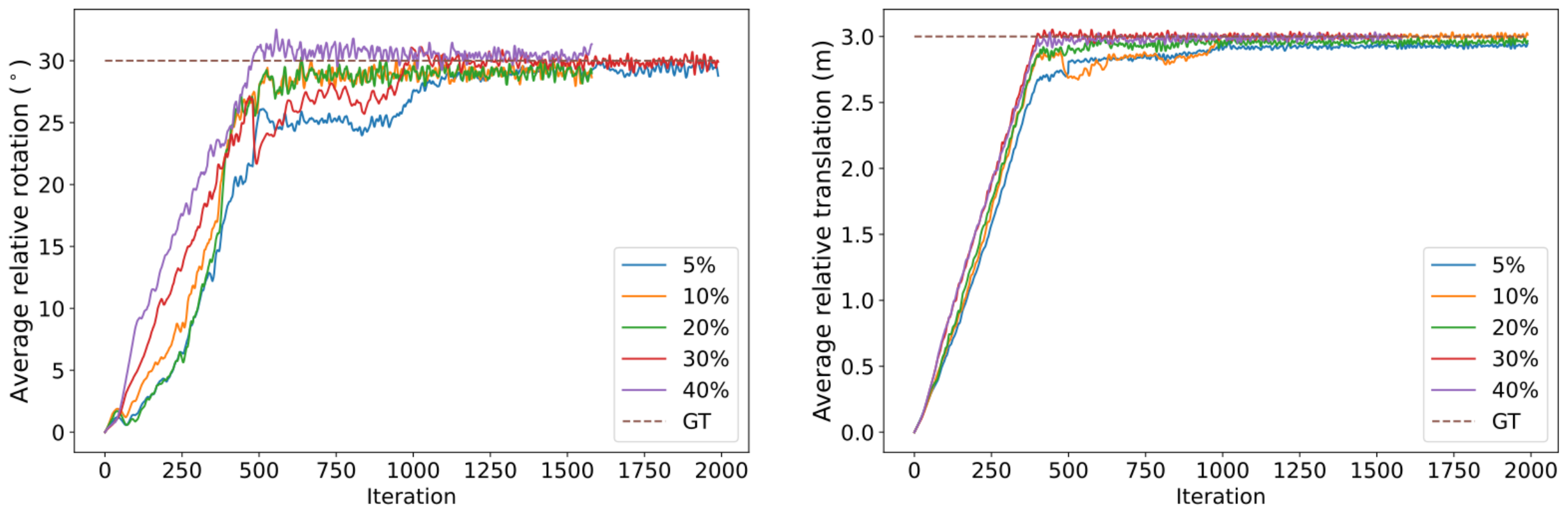}
         \vspace{-2.2 em}
    \caption{\textcolor{black}{Optimizing the relative poses for two robots with varying overlaps.} 
    }
    \label{fig:varying_overlaps}
    \vspace{-1.2 em}
\end{figure}
\subsection{Trajectory Overlap Analysis}
\label{sec:C}
In this experiment, the sensitivity of the convergence of the relative poses with respect to the overlaps between the views of the robots is analyzed. 
For this experiment, two robots are considered. There are no common images between the robots, and the overlaps are only in the trajectories. Fig.~\ref{fig:chair_overlaps} shows the top-down view of the overlaps, ranging from 5\% to 40\%, i.e.  [5, 10, 20, 30, 40]\%. 
The overlaps are determined using $x$ and $y$ coordinates of camera poses, assuming the poses are known in a coordinate frame. To divide the views among $N$ robots, first, the $x-y$ plane containing the desired object (the Chair, in this case) is divided into $N$ sectors, centered on the object. This will generate a 0\% overlap for $N$ robots, which then is extended to increase the overlap, as shown in Fig.~\ref{fig:chair_overlaps}. 

For two robots, once the overlaps are determined, the poses of robot $R^2$ are manipulated to create a relative displacement with respect to $R^1$, by translating them for \textcolor{black}{3m} along each axis and rotating them for 30$^\circ$ around each axis.  
Then optimization is performed, assuming an initial estimate for the relative pose with \textcolor{black}{zero} for each translation component and each rotation.

Fig.~\ref{fig:varying_overlaps} shows the convergence results for different overlaps, for 2000 steps of Di-NeRF training. The ground-truth relative poses are marked with GT. For each overlap, the plot shows the average estimates of translation and rotation components. The training configurations are the same for all overlaps. It is observed that large overlaps improve the convergence. 
\subsection{Waymo dataset - Unbounded Scenes}
\label{sec:D}

\begin{figure}[t!]
\vspace{-1 em}
    \centering
    \includegraphics[width=1\linewidth]{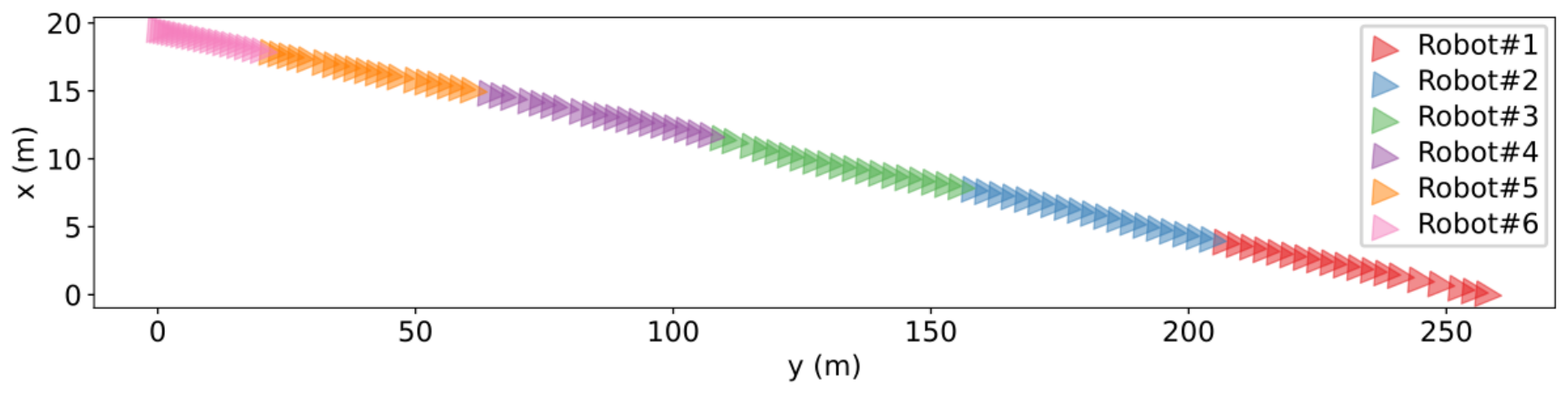}
    \vspace{-2.2 em}
    \caption{A sequence of a Waymo trajectory is divided into six segments, i.e. robots, and used to evaluate Di-NeRF.}
    \label{fig:waymo_trajectory}
    \vspace{1 em}
    \centering
    \color{black}
    \includegraphics[width=1\linewidth]{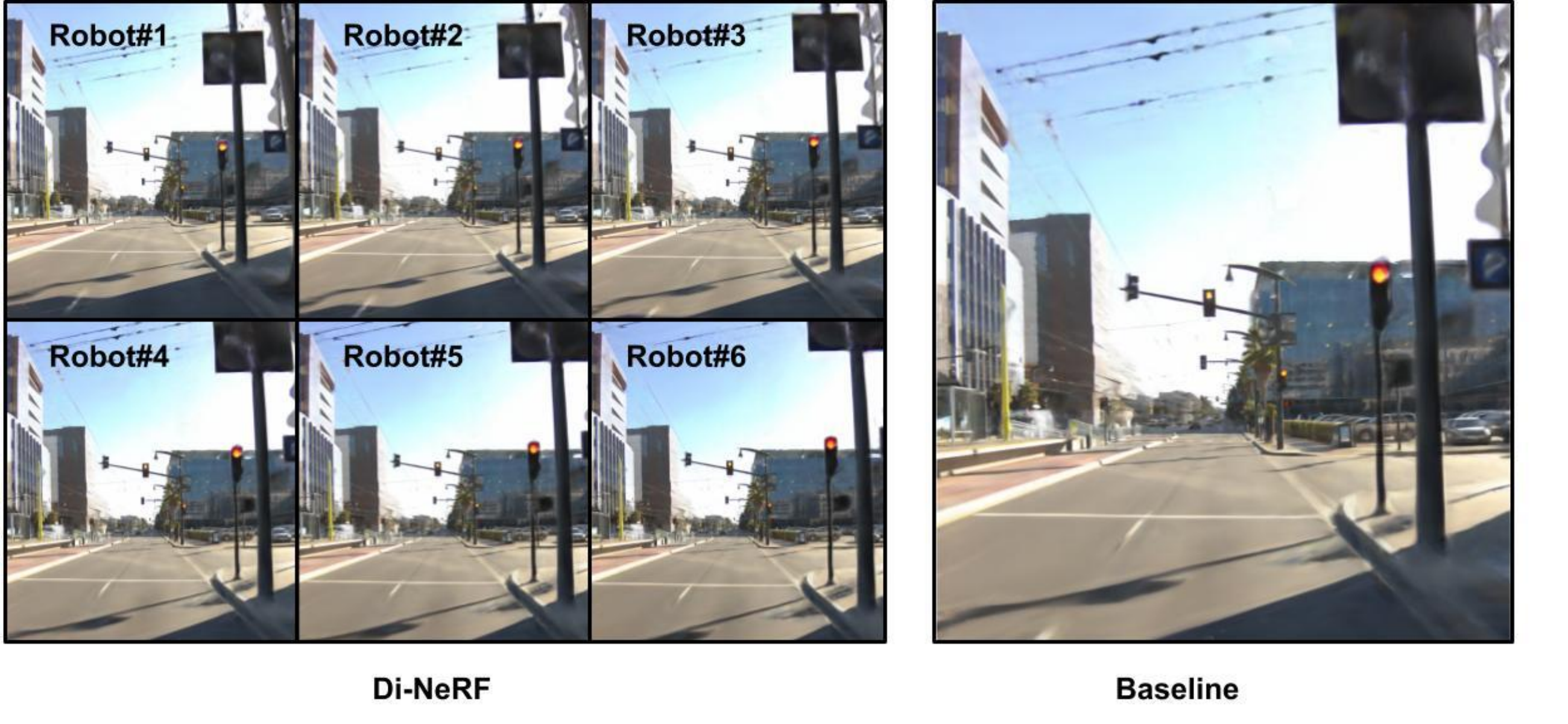}
    \vspace{-2.2 em}
    \caption{\textcolor{black}{Rendered images from Di-NeRF on the Waymo dataset, all six robots demonstrated the capability to render similar images with quality comparable to that achieved through centralized processing. }}
    \label{fig:waymo_renders}
    \vspace{-1. em}
\end{figure}

The San Francisco Mission Bay dataset (Waymo)~\cite{tancik2022blocknerf} captures 12,000 images over 100 seconds along a 1.08 km route, using 12 cameras mounted on a car in an urban environment with dynamic objects and reflective surfaces. A 286m data segment with 233 images from a single roof-mounted camera providing a full surround view was selected for this experiment (Fig.~\ref{fig:waymo_trajectory}). The sequence was divided into six segments to resemble a multi-agent setup. Initial relative poses are offset by 1m from actual values. For larger offsets, the algorithm fails due to lack of overlap.

Fig.~\ref{fig:waymo_renders} shows the rendering result for the \textcolor{black}{baseline} and Di-NeRF. All robots can render a view that they have never seen directly. The average values of PSNR and SSIM are \textcolor{black}{25.60} and \textcolor{black}{0.831} over 6 robots and 25.39 and 0.848 for the baseline. Table~\ref{table4} shows detailed results. \textcolor{black}{We divided the Waymo dataset into segments to benchmark Di-NeRF against Block-NeRF~\cite{tancik2022blocknerf}. Table~\ref{table5} shows the results. The objective of this experiment is not to surpass Block-NeRF~\cite{tancik2022blocknerf}, which is centralized. The goal is to achieve comparable results between distributed and centralized systems.}

\subsection{\textcolor{black}{Convergence of the Relative Poses}}
\label{sec:E}
In this section, convergence analysis with different initial relative pose values for the Barn sequence is provided. For simplicity, two robots are considered. The actual relative translation is 3m in all directions (i.e. x, y, and z), and the average initial values for three translations start from 0m to 2m (off by -1m to 1m), with increments of 0.02m. Similarly, the rotation ground truth is 45$^\circ$, and the initial value ranges from 0$^\circ$ to 90$^\circ$ (off by -45$^\circ$ to 45$^\circ$), with increments of 0.9$^\circ$. 
The translation error averages (0.86, 0.76)cm, and the rotation error (0.32, 0.38)$^\circ$, as shown by the blue curve in Fig.~\ref{fig:abalation}. Experimentally, the algorithm converges even with initial values far from the ground truth.

\begin{figure}[t!]

    \centering
    \color{black}
    \includegraphics[width=1\linewidth]{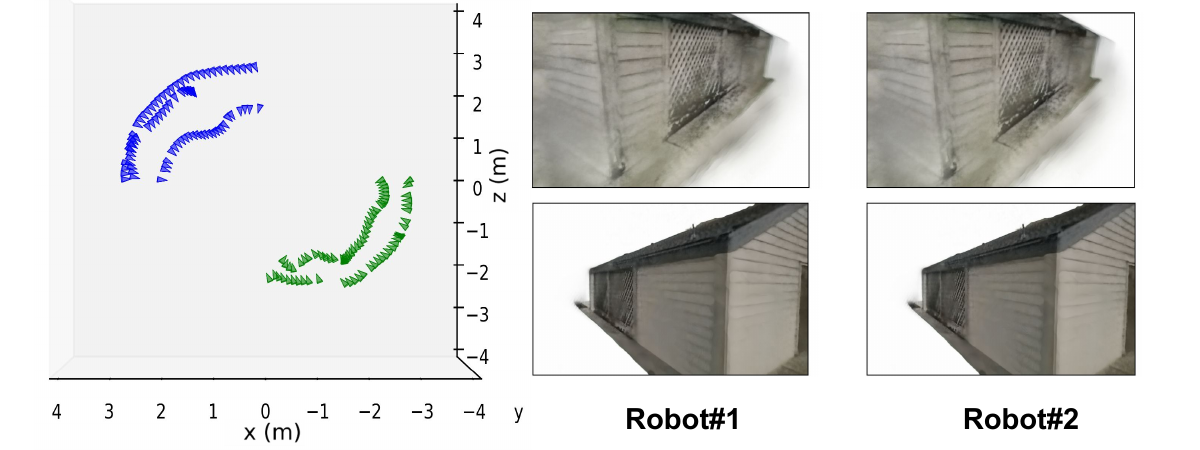}
    \vspace{-2.2 em}
    \caption{\textcolor{black}{({Left}) Trajectories of two robots from the Barn sequence showing no overlaps in views. Rendered views after Di-NeRF training: ({Top Right}) with unknown relative pose; ({Bottom Right}) with a known relative pose}.}
    \label{fig:no_overlap}
    \vspace{1 em}
    \centering
    \color{black}
    \includegraphics[width=1\linewidth]{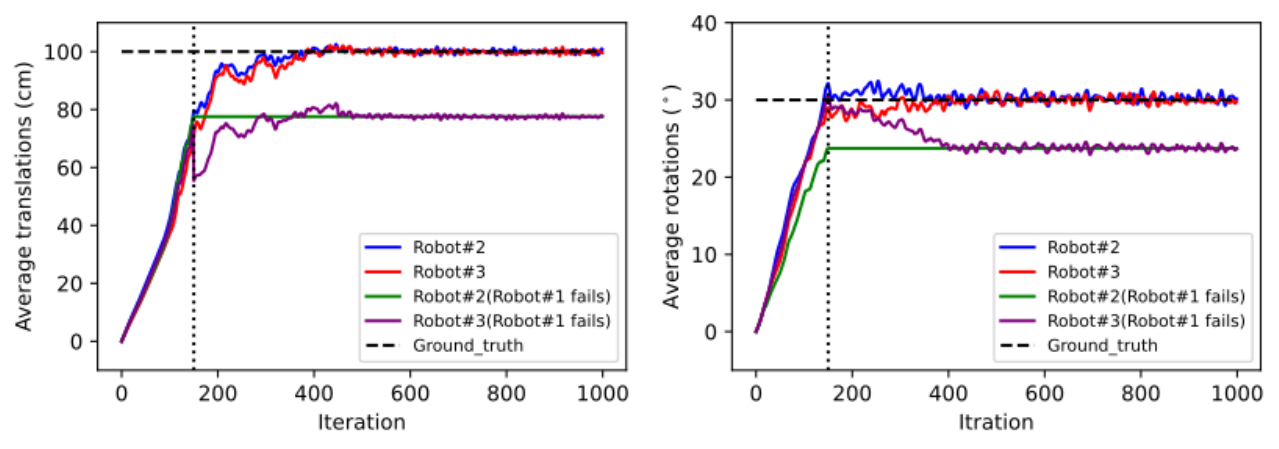}
    \vspace{-2.2 em}
    \caption{\textcolor{black}{Average values for relative rotations and translations for $R^2$ and $R^3$, in two cases: $R^1$ doesn't fail (red and blue lines) and $R^1$ fails (green and purple lines) after 150 iterations. When $R^1$ fails, the global coordinate is transferred to $R^2$, and $R^3$ is optimizing for relative pose with respect to $R^2$.}}
    \label{fig:ref_failure}
     \vspace{-1.2 em}
\end{figure}
\subsection{\textcolor{black}{No Overlap Setup}}
\label{sec:F}
\textcolor{black}{Even when there is no direct path overlap, scenes may be viewed from varying angles, allowing for relative pose calculation. For scenarios with no view overlaps, we have included an experiment using the Barn sequence. Fig.~\ref{fig:no_overlap} depicts the trajectories for two robots with results. Di-NeRF struggles to compute relative poses with no view overlaps. However, knowing the relative poses, Di-NeRF merges the two models in a distributed fashion despite no absolute overlap.}
\subsection{\textcolor{black}{Reference Robot Failure}} 
\label{sec:refrence}
\textcolor{black}{If the robot chosen as the global coordinate frame fails or does not communicate, the global frame is switched to another robot. Fig.~\ref{fig:ref_failure} shows an experiment where after 150 iterations, the reference robot is changed. Di-NeRF successfully optimizes the relative pose with respect to a new origin.}
\subsection{Technical Details of the Experiments} 
\label{sec:details}

In this paper, three datasets are used to evaluate Di-NeRF: 
the synthetic dataset~\cite{mildenhall2021nerf} (800$\times$800 pixels), 
the Tanks and Temples~\cite{Knapitsch2017} (1920$\times$1080 pixels), and 
the Waymo dataset~\cite{tancik2022blocknerf} (1217$\times$1096 pixels). 
For all datasets, the batch size is 2048, the learning rate is 0.01, \textcolor{black}{the weight for depth loss $\gamma$ in Eq.~\eqref{depth_loss} is 0.01} and the value of $\rho$ in Eq.~\eqref{eq-C-ADMM} is 0.001. $\rho$ is the weight for the quadratic term and is the step size in the gradient ascent for the dual variable optimization.
The number of steps before a communication round is \textcolor{black}{200. There are 5 communication rounds in all experiments.}

\noindent \textbf{Hardware and Software Configuration}: The training was conducted on one NVIDIA RTX A5000 GPU with 24 GB of memory. We utilized PyTorch as our deep learning framework.

\noindent \textbf{NeRF Network Configuration}: The network $\theta$ has 8 layers with 256 channels each. The hashmap size \textcolor{black}{is $2^{15}$ for synthetic sequences and $2^{19}$ for the Barn and Waymo sequences.} 
Table~\ref{table6} shows the training times for a fully connected graph. 
\textcolor{black}{The memory required for a NeRF model is shown in Table~\ref{table7}. A robot needs 40.7 MB to exchange models, which can be reduced by compressing them when bandwidth is limited.}
\begin{table}[t!]
\vspace{-0.5 em}
    \centering 
    \caption{Di-NeRF and baseline (Bsln.) performances for the Mission Bay sequence for 6 robots.
    $R\#1$ is the reference frame.}
    \vspace{-1 em}
    \label{table4}
    \begin{tabular}{|c|c|c|c|c|c|c||c|}
    \hline
    &R\#1&R\#2&R\#3&R\#4&R\#5&R\#6&Bsln.\\
    \hline
    PSNR$ \uparrow$&\textcolor{black}{25.65}&\textcolor{black}{25.24}&\textcolor{black}{25.56}&\textcolor{black}{25.67}&\textcolor{black}{25.78}&\textcolor{black}{25.70}&25.39\\
    \hline
    SSIM $\uparrow$&\textcolor{black}{0.826}&\textcolor{black}{0.854}&\textcolor{black}{0.815}&\textcolor{black}{0.819}&\textcolor{black}{0.838}&\textcolor{black}{0.839}&0.848\\
    \hline
    $\delta R^{\circ}\downarrow$&-&\textcolor{black}{2.89}&\textcolor{black}{2.86}&\textcolor{black}{2.81}&\textcolor{black}{2.91}&\textcolor{black}{2.79}&-\\
    \hline
    \hspace{-.5 em}$\delta t${\scriptsize (cm)}$\downarrow$&-&\textcolor{black}{4.2}&\textcolor{black}{4.0}&\textcolor{black}{4.5}&\textcolor{black}{4.2}&\textcolor{black}{4.7}&-\\
    \hline
    \end{tabular}
    \vspace{1 em}
    \centering 
    \color{black}
    \caption{Di-NeRF and Block-NeRF, on the Mission Bay sequence.}
    \vspace{-1 em}
    \label{table5}
    \begin{tabular}{|c|c|c||c|c|c|}
    \hline
    \multicolumn{3}{|c||}{Block-NeRF (centralized)}&\multicolumn{3}{|c|}{Di-NeRF (distributed)}\\
    \hline
    No. Blocks&PSNR&SSIM&No. Agents&PSNR&SSIM\\
    \hline
    2	&\textcolor{black}{\textbf{25.65}}	&\textcolor{black}{\textbf{0.871}}	&2	&25.22	&0.866\\
    \hline
    8	&\textcolor{black}{24.56}	&\textcolor{black}{\textbf{0.892}}	&8	&\textbf{25.41}	&0.854\\
    \hline
    16	&\textcolor{black}{\textbf{27.1}}	&\textcolor{black}{\textbf{0.903}}	&16	&25.34	&0.861\\
    \hline
    \end{tabular}
    \vspace{-1 em}
\end{table}

\begin{table}[t!]
  \color{black}
  \centering
  \caption{Di-NeRF training time comparison (sec/itr/robot).}
  \vspace{-1 em}
  \label{table6}
  \footnotesize
  \setlength{\tabcolsep}{3pt}
  \renewcommand{\arraystretch}{1.2}
  \begin{tabular}{|c|c|c|c|c|}
    \hline
    \textbf{Dataset} & \textbf{\#Robots} & \textbf{Di-NeRF} & \textbf{\textcolor{black}{Di-NeRF;known $T_i^g$}}& \textbf{\textcolor{black}{Baseline}}  \\
    \hline
    Synthetic Dataset & 2 & 4.27 & \textcolor{black}{0.98} & 0.94  \\
    \hline
    Tank and Temple & 5 & 4.78 & \textcolor{black}{0.98} & 0.93  \\
    \hline
    Mission Bay & 6 & 4.73 & \textcolor{black}{0.97} & 0.95  \\
    \hline
  \end{tabular}
\vspace{1 em}
\color{black}
    \centering 
    \caption{Images and NeRF size with the required bandwidth}
    \vspace{-1 em}
    \label{table7}
    \begin{tabular}{|c|c|c|c|}
    \hline
    Sequence&Seq. Size (MB)&Model Size (MB)&BW per Link\\
    \hline
    Chair&31.04&4.07&40.7\\
    \hline
    Lego&42.9&4.07&40.7\\
    \hline
    Hotdog&40.9&4.07&40.7\\
    \hline
    Barn&481.5&64.07&640.7\\
    \hline
    Waymo&428.2&64.07&640.7\\
    \hline
    \end{tabular}
\color{black}
\vspace{-1 em}
\end{table}
\begin{figure}[t!]
    \centering
    \color{black}
    \includegraphics[width=1\linewidth]{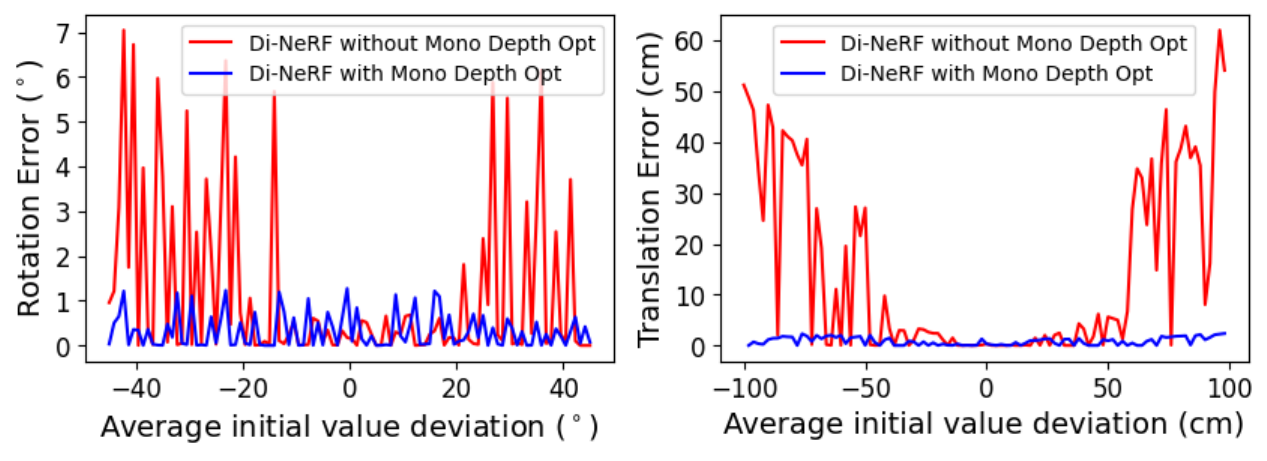}
    \vspace{-2.2 em}
    \caption{\textcolor{black}{Impact of initial translation and rotation values on relative pose convergence (the Barn sequence) and performance differences between two approaches in optimizing the Di-NeRF model. The initial translation deviation from ground truth ranges from -1m to 1m, and for rotation from -45$^\circ$ to 45$^\circ$. The first approach (shown in blue) integrates monocular depth estimation into the relative pose optimization, while the second approach (depicted in red) does not incorporate monocular depth estimation.}}
    \label{fig:abalation}
 \vspace{-1.2 em}
\end{figure}
\subsection{\textcolor{black}{Ablation Study}}
\label{sec:abalation}
\textcolor{black}{We conducted an ablation study to examine the effects of incorporating monocular depth estimation into relative pose optimization. The findings are detailed in Fig~\ref{fig:abalation}. While joint optimization of NeRF and relative pose often faces convergence issues with inaccurate initial position error estimates, the use of monocular depth estimation significantly mitigates this problem by correcting depth distortions. }

\section{Conclusion and Future Work} \label{sec:conclusion}
We introduced Di-NeRF, the first fully distributed NeRF used for multi-robot systems and 3D scenes. 
We demonstrate its versatility across synthetic and real datasets. Di-NeRF harnesses the advantages of NeRF in distributed learning, relying on RGB input for relative pose optimization. Our analyses reveal its applicability to varying numbers of robots, diverse connection types, and distinct trajectory overlaps.
    
In this work, the relative camera extrinsics are optimized. Optimizing the intrinsics is useful when cameras on the robots have different settings. Also, the base NeRF model for each robot has limitations in unbounded scenes. A model, designed for unbounded scenes, can be integrated into this project. Finally, we assume the local camera poses are known but with different origins. By jointly optimizing NeRF and local camera poses, one can eliminate this requirement. 
\textcolor{black}{Also, by optimizing the architecture of NeRF, Di-NeRF can be efficient in bandwidth use compared to directly sending RGB images. This demonstrates Di-NeRF's advantage in distributed systems, where communication is localized, reducing bandwidth needs unlike in centralized systems that require comprehensive data transmission. The core strength of Di-NeRF lies in facilitating collaborative mapping in challenging environments, emphasizing the distributed nature of learning and system scalability where broad communication may not be feasible.}

\bibliographystyle{IEEEtran}
\bibliography{./bibliography/all}
\section{Supplementary Material}
\subsection{The Auxiliary parameter in Di-NeRF}
Considering the distributed optimization in Di-NeRF that is subject to the constraint $\theta_i=z_{ij}$, the following sub-problem optimization can be written for each robot as follows using the auxiliary variable $z_{ij}$: 

\begin{align}
    \textcolor{black}{{\Gamma}}_i^{(k+1)} &= \underset{\textcolor{black}{{\Gamma}}}{\operatorname{argmin}} \\
    &\quad \{L_i\left(\textcolor{black}{{\Gamma}}_i\right) + \theta_i^{\top} y_i^{(k)} + \rho \sum_{j \in \mathcal{N}_i}\left\|\theta_i - \bar{\theta}^{(k)}\right\|_2^2\}
    \label{eq1}
\end{align}
\noindent where $L_i$ is the local loss function for each robot, and ${\Gamma}_i = \{\theta_i, T_i^g,\boldsymbol{\psi}_i\}$). \(\theta_i\) is the network parameters of the robot \(i\) which is communicating with the robot \(j\). The parameter \(\rho\) is the weight for the quadratic term $\sum_{j \in \mathcal{N}_i}\left\|\theta_i-z_{ij}\right\|_2^2$ and the step size in the gradient ascent of the dual variable $y_i$. $\mathcal{N}_i$ is the set of neighbours of robot $i$, and $T^g_i$ is the relative pose of robot $i$ with respect to the global coordinate.
According to \eqref{eq1}, the local loss function for each robot can be updated according to the following \cite{boyd2011distributed}: 

\begin{equation}
\begin{aligned}
\mathcal{L}_i(\textcolor{black}{{\Gamma}_i}) = L_i(\textcolor{black}{{\Gamma}_i})+ \theta_i^{\top} y_i + \frac{\rho}{2} \sum_{j \in \mathcal{N}_i }{\left\| \theta_i-z_{ij}\right\|_2^2} \ ,
\end{aligned}
\label{eq2}
\end{equation}

\noindent where $\mathcal{L}_i$ is the updated loss function for each robot based on Di-NeRF. The dual variable $y_i$ update according to C-ADMM is as follows:
\begin{equation}
y_i^{(k+1)} = y_i^{(k)} + \rho (\theta_i^{(k+1)}-z_{ij}^{(k+1)}),
\label{eq3}
\end{equation}
To update the auxiliary parameter $z_{ij}$ given $\theta_i-z_{ij}=0$ and \eqref{eq1}, the z update subproblem is as follows:
\begin{equation}
\min_{z^{(k+1)}} \quad -\sum_{i\in \mathcal{N}_i} y_i^{(k)} z^{(k+1)} + \frac{1}{2} \sum_{i\in \mathcal{N}_i} \left\| \theta_i^{(k+1)} - z^{(k+1)} \right\|^2
\label{eq4}
\end{equation}
\noindent By taking the first derivative from \eqref{eq4}, the following equation can be driven for $z_{ij}$: 
\begin{equation}
z^{(k+1)} = \frac{1}{\mathcal{N}}\sum_{i\in \mathcal{N}_i}{(\theta_i^{(k+1)} + \frac{1}{\rho} y_i^{(k)})}
\label{eq5}
\end{equation}
\noindent Dual variable update is an equality, not an optimization problem and is called a central collector or fusion center. Equation \eqref{eq5} can be simplified further by writing it as:
\begin{equation}
z^{(k+1)} = \Bar{\theta}^{(k+1)} + \frac{1}{\rho} \Bar{y}^{(k)}
\label{eq6}
\end{equation}
\noindent Substitution of \eqref{eq6} into the average value of $y_i$ over $\mathcal{N}_i$ in \eqref{eq3} (i.e. $\Bar{y}_i^{(k+1)} = \Bar{y}_i^{(k)} + \rho (\Bar{\theta}_i^{(k+1)}-z^{(k+1)})$), yields $\Bar{y}_i^{(k+1)} = 0$. Further substitution of this result into \eqref{eq6} leads to the following final equation: 

\begin{equation}
z_{ij}^{(k+1)} = \frac{1}{N}\sum\theta_i^{(k)} \coloneqq \bar{\theta}^{(k)}~,
\end{equation}

During each iteration, $k$, every robot, indexed by $i$, independently solves its own subproblem to determine the value of the global variable  $\theta_i^{(k)}$. The robot incurs a penalty proportional to the deviation of its variable from the mean value of the global variable, as computed from all robots in the preceding iteration.

\end{document}